# Mapping Methane - The Impact of Dairy Farm Practices on Emissions Through Satellite Data and Machine Learning

**Hanqing Bi [1,2] and Suresh Neethirajan [2,3]***
[1]Faculty of Computer Science, 6050 University Avenue, Dalhousie University, Halifax, Canada
[2]Faculty of Mathematics, 200 University Ave W, Waterloo, Ontario, Canada
[3]Faculty of Agriculture, Agricultural Campus, PO Box 550, Dalhousie University, Truro, NS, B2N 5E3
*Correspondence: sneethir@gmail.com

**Abstract:** This study investigates the correlation between dairy farm characteristics and methane concentrations as derived from satellite observations in Eastern Canada. Utilizing data from 11 dairy farms collected between January 2020 and December 2022, we integrated Sentinel-5P satellite methane data with critical farm-level attributes, including herd genetics, feeding practices, and management strategies. Initial analyses revealed significant correlations with methane concentrations, leading to the application of Variance Inflation Factor (VIF) and Principal Component Analysis (PCA) to address multicollinearity and enhance model stability. Subsequently, machine learning models—specifically Random Forest and Neural Networks—were employed to evaluate feature importance and predict methane emissions. Our findings indicate a strong negative correlation between the Estimated Breeding Value (EBV) for protein percentage and methane concentrations, suggesting that genetic selection for higher milk protein content could be an effective strategy for emissions reduction. The integration of atmospheric transport models with satellite data further refined our emission estimates, significantly enhancing accuracy and spatial resolution. This research underscores the potential of advanced satellite monitoring, machine learning techniques, and atmospheric modeling in improving methane emission assessments within the dairy sector. It emphasizes the critical role of farm-specific characteristics in developing effective mitigation strategies. Future investigations should focus on expanding the dataset and incorporating inversion modeling for more precise emission quantification. Balancing ecological impacts with economic viability will be essential for fostering sustainable dairy farming practices.

---

## 1. Introduction

The escalating threat of climate change has propelled greenhouse gas emissions to the forefront of global environmental concerns. Among these, methane has emerged as a particularly potent contributor to global warming, with its impact significantly exceeding that of carbon dioxide on a per-molecule basis [19, 41]. The urgency to mitigate methane emissions has intensified, with atmospheric concentrations rising alarmingly, reflecting a fourfold increase in global emissions over recent decades [33, 50]. This surge poses a critical challenge to international climate goals, particularly those outlined in the Paris Agreement.

The dairy sector is identified as a substantial source of anthropogenic methane emissions, significantly contributing to the agricultural sector's greenhouse gas footprint [2, 48]. The impact of methane from dairy production is particularly concerning due to its global warming potential—approximately 28 times that of carbon dioxide over a 100-year period [42, 12]. This necessitates targeted mitigation strategies within the industry. Primary methane sources in dairy farming include enteric fermentation and manure management practices [15, 43, 47]. The biological complexities underlying these emissions present both challenges and opportunities for mitigation. Understanding and addressing these sources are essential not only for climate change mitigation but also for promoting sustainable agricultural practices that meet global food demands while minimizing environmental impact.

The intricate relationship between dairy farm characteristics and methane emissions demands a multifaceted mitigation approach. Factors such as herd genetics, feeding strategies, and management practices significantly influence methane output, yet the relative importance and interactions among these factors remain poorly understood, hindering the development of effective, targeted mitigation strategies. Recent advancements in satellite technology and data analytics have opened new avenues for monitoring and quantifying methane emissions at unprecedented scales. The launch of the Sentinel-5P satellite, equipped with the Tropospheric Monitoring Instrument (TROPOMI), has revolutionized our capacity to detect and measure atmospheric methane concentrations with high spatial and temporal resolution [17, 45]. This technology offers a unique opportunity to bridge the gap between farm-level practices and regional methane concentrations, potentially unveiling patterns previously obscured by the limitations of ground-based measurements.

The integration of satellite data with ground-based measurements and atmospheric transport models represents a significant advancement in creating comprehensive methane emission estimates [50, 51]. This approach enhances our understanding of methane dynamics across various spatial scales and provides a powerful tool for validating and refining emission inventories. The capability to detect and quantify emissions from space offers unprecedented potential for identifying and addressing methane hotspots, particularly in dairy farming regions. The advent of advanced machine learning techniques further amplifies the potential of satellite-based methane monitoring. These sophisticated algorithms can analyze vast datasets to identify patterns and anomalies that may indicate emission sources or trends [40, 24]. Applying machine learning to satellite data holds promise for enhancing the detection and quantification of methane emissions, potentially revealing insights that traditional analytical methods might overlook.

Inversion modeling techniques have emerged as vital tools in translating satellite-observed methane concentrations into actual emission estimates [23]. This methodology allows researchers to infer emission strengths and locations from atmospheric measurements, providing valuable insights into the relationship between farm practices and regional methane concentrations. The synergy between multiple satellite platforms, such as Sentinel-5P and GOSAT, could further enhance monitoring capabilities [4, 44]. Continuous satellite monitoring enables the detection of long-term methane emission trends, essential for assessing the effectiveness of mitigation strategies and policies. This capability is particularly relevant for the dairy sector, where changes in farming practices may take time to manifest in observable emission reductions. Long-term satellite observations can provide critical data for evaluating various interventions and guiding future policy decisions.

The rise of artificial intelligence and machine learning offers new possibilities for analyzing emission structures and optimizing mitigation strategies in the dairy industry. By assigning varying importance to different farm characteristics, these technologies can potentially identify innovative solutions for reducing methane emissions from dairy farms [35, 27]. This data-driven approach could lead to more targeted and effective interventions tailored to specific conditions of individual farms or regions. Despite these advancements, challenges persist in accurately attributing methane emissions to specific sources and in developing effective mitigation strategies for the dairy sector. The complex interplay among farm management practices, animal genetics, and environmental factors complicates efforts to pinpoint the most impactful interventions. Moreover, potential trade-offs between methane reduction strategies and other aspects of dairy sustainability—such as animal welfare and economic viability—must be carefully considered.

This study aims to address these challenges by exploring the correlation between dairy farm factors and satellite-derived methane concentrations in Eastern Canada. By leveraging advanced data analytics and machine learning techniques, we seek to uncover insights that can inform breeding programs, management practices, and policy decisions aimed at reducing the dairy sector's environmental impact. Our approach involves a comprehensive analysis of various farm characteristics, including herd genetics, feeding practices, and management strategies, to identify the most significant factors influencing methane emissions.

## 2. Related Work

Recent advancements in satellite technology and data analytics have revolutionized the monitoring and analysis of methane emissions from the dairy sector. The Mooanalytica research group at Dalhousie University has made notable contributions in this field, leveraging satellite data to assess methane emissions from Canadian dairy farms and processors.

### 2.1. Satellite-Based Methane Monitoring

Satellite data collection has emerged as a reliable and cost-effective method for gathering large-scale environmental data [20, 21]. The Sentinel-5P satellite, equipped with TROPOMI, has proven particularly instrumental in this regard, accurately collecting data on various atmospheric gases, including methane, through advanced spectroscopy methods [45]. Orbiting at an altitude of 824 km, Sentinel-5P provides daily global coverage with a spatial resolution of 7 km × 7 km, ensuring consistent data collection [28, 38]. This technology has been applied in numerous studies, particularly in greenhouse gas emissions research, with recent efforts focusing on integrating artificial intelligence for enhanced analysis [22, 25].

### 2.2. Application to the Canadian Dairy Sector

Recent studies utilizing Sentinel-5P data have analyzed methane emissions from over 575 dairy farms and 384 dairy processors across Canada over an eight-year period, revealing significant seasonal and provincial variations in emissions, with autumn emissions peaking and Ontario exhibiting the highest overall emissions [7] (Bi and Neethirajan, 2024). Building on this work, an AI-driven benchmarking tool for emission reduction in Canadian dairy farms has been developed, integrating satellite-derived methane emission data with advanced machine learning technologies and geospatial analysis [5, 10, 32].

### 2.3 Advanced Analytical Techniques

The integration of satellite data with machine learning models has significantly enhanced the detection and quantification of methane emissions. Studies have demonstrated how advanced machine learning techniques applied to satellite data improve the identification of emission patterns and anomalies [40, 24]. Inversion modeling techniques have been crucial in translating satellite-observed methane concentrations into actual emission estimates [23], while the synergy between multiple satellite platforms has further bolstered methane monitoring capabilities [3, 44].

### 2.4 Artificial Intelligence in Emission Analysis

The rise of artificial intelligence has opened new avenues for emission structure analysis in the dairy industry [26]. Recent studies have demonstrated how AI can assign varying importance to different farm characteristics, potentially identifying novel solutions for reducing methane emissions [35, 18, 27]. The Mooanalytica group's work exemplifies this approach, providing actionable insights for implementing effective emission reduction strategies.

2.5 Challenges and Future Directions

Despite these advancements, challenges remain in accurately attributing methane emissions to specific sources and in developing effective mitigation strategies. The complex interplay between farm management practices, animal genetics, and environmental factors complicates efforts to identify the most impactful interventions [37].

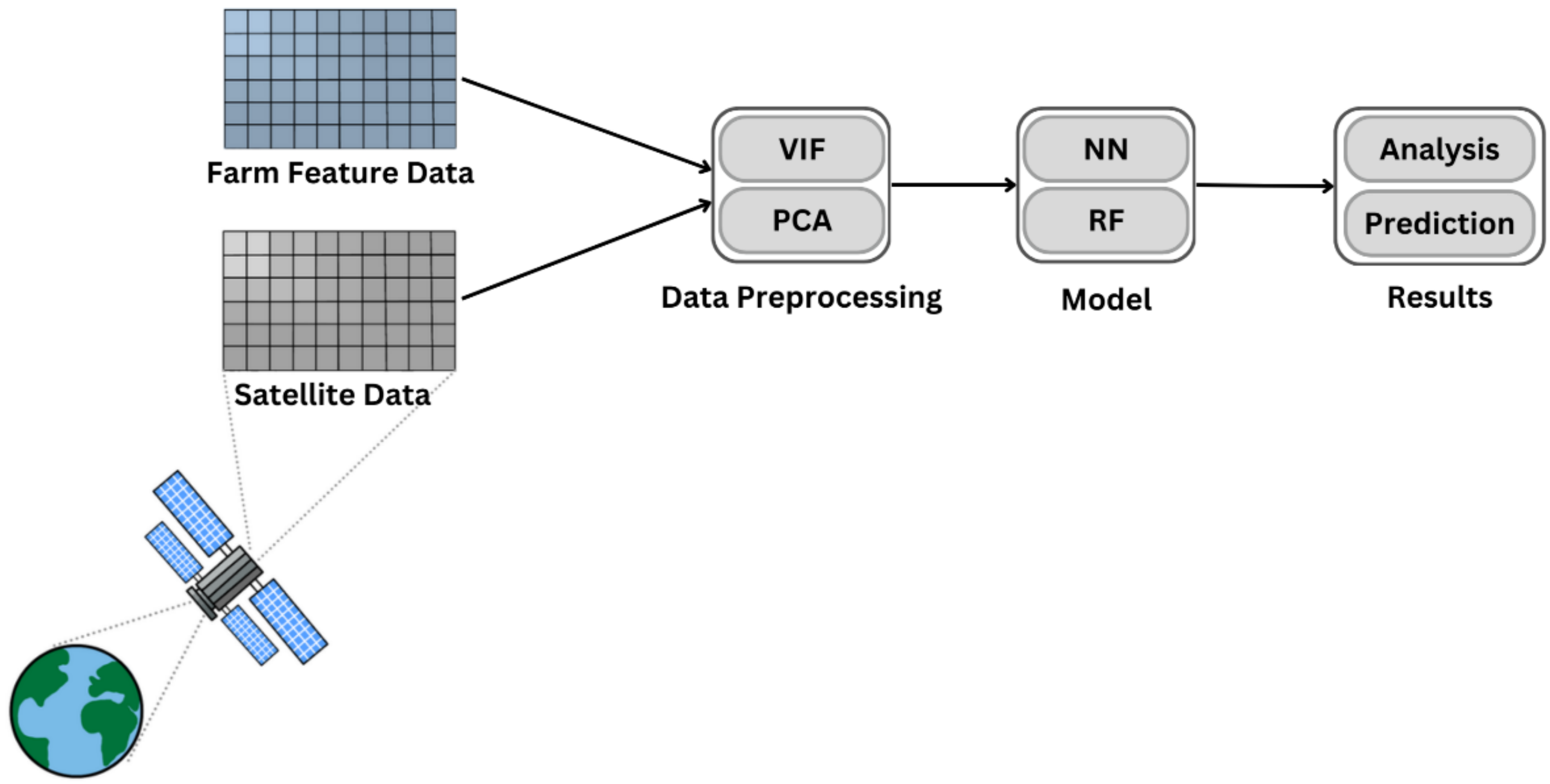

**Figure 1.** Schematic representation of the data analysis process, including data collection, feature analysis, and machine learning modeling to assess the relationship between dairy farm characteristics and methane emissions.

## 3. Materials and Methods

3.1 Data Collection and Integration

Data were collected from 11 dairy farms located in Atlantic Canada, specifically in Nova Scotia, Prince Edward Island, and New Brunswick. This dataset was obtained through Lactanet, which provided comprehensive insights into various operational aspects of these farms. Although the data were not collected simultaneously across all farms, the information gathered offered valuable insights into the factors influencing methane emissions. A total of 36 factors related to dairy farm operations were identified and categorized into four main groups: basic information, production performance, reproduction performance, culling performance, and genetics information. Basic information included the Province code, Lactanet herd ID, and the date the herd began milk recording. Production performance metrics encompassed average milk yield at test date (kg/day), average milk fat yield at test date (kg/day), average milk protein yield at test date (kg/day), average somatic cell count (SCC) at test date (x 1000 cells/ml), and average milk urea nitrogen at test date (mg/dl). Additionally, MTP predicted rolling 12-month herd averages for 305-day milk yield, fat yield, and protein yield were included. Reproduction performance factors comprised rolling 12-month herd averages for gestation length (days), dry period length (days), days open (days), calving interval (days), days to first service (days), and total number of breeding services per year. Culling performance data included percentages of cows culled for voluntary reasons in the last 12 months, involuntary reasons, cows left for various reasons, mortality rates, and percentages sold or left for reproduction or health issues. Genetics information encompassed the count of animals in the herd with a genetic evaluation, EBVs for milk yield, fat yield, protein yield, fat percentage, protein percentage, relative breeding value for SCC linear score, lifetime performance index, and Pro$ index. To complement the farm-level data, we utilized Sentinel-5P satellite data to obtain methane column volume mixing ratio dry air bias-corrected measurements corresponding to each farm's location during the data collection period. This integration allowed us to analyze methane emissions in relation to specific farm characteristics.

### 3.2 Data Collection Time Frame and Temporal Alignment

Farm-level data were collected from January 2020 to December 2022, spanning a total duration of three years. Data collection occurred monthly to capture seasonal variations that could affect methane emissions, such as changes in feed composition or herd management practices. This frequency facilitated the assessment of temporal patterns and potential seasonal effects on methane production. To align farm data with satellite observations from Sentinel-5P, we synchronized farm data collection dates with the satellite's overpass times. Sentinel-5P orbits the Earth at an altitude of 824 km with a 100-minute cycle, enabling daily data collection over the same location with a spatial resolution of $7 \times 7$ km². Methane concentration data were extracted for the geographic coordinates of each farm on corresponding dates. When multiple satellite observations were available for a single farm data point, we averaged the methane concentration values to derive representative measurements. Potential errors or biases in matching ground-level farm data with satellite-derived methane concentrations were considered. Factors such as atmospheric conditions, cloud cover, and the satellite's spatial resolution could affect data accuracy. To mitigate these issues, satellite data collected under unfavorable atmospheric conditions were excluded based on quality flags provided in the Sentinel-5P dataset. Additionally, we cross-referenced farm locations with land use data to identify and account for other potential methane sources within the satellite's footprint.

### 3.3 Data Processing and Analysis

Data processing was a critical step in transforming raw information into actionable insights. The farm-level and satellite data were organized and cleaned using Python scripts to prepare them for subsequent analyses. Specifically, preprocessing included the handling of missing data, outlier detection, and normalization. Given the large number of features (36 factors related to farm operations), feature engineering was performed using Variance Inflation Factor (VIF) and Principal Component Analysis (PCA) to address multicollinearity and reduce dimensionality, respectively. VIF was applied to identify and remove features with high multicollinearity. This step was crucial for improving the stability and interpretability of the subsequent models, especially when training Random Forest. Meanwhile, PCA was used as a dimensionality reduction tool to summarize the data into fewer components that still captured most of the variance, which helped in reducing the complexity of Neural Network training. These preprocessing steps provided a robust basis for machine learning analysis, enabling better performance and interpretability [9].

### 3.4 Emissions Profiling and Trend Analysis

We employed two machine learning models to predict methane emissions: Random Forest (RF) and Neural Network (NN). Both models were trained using the preprocessed dataset, with PCA-transformed data being used primarily for NN to optimize its training efficiency by reducing input dimensionality. RF, in contrast, was more effective when trained on the dataset after handling multicollinearity via VIF, preserving interpretability of feature contributions. The Random Forest model was chosen due to its capability to handle a large number of features and its robustness against overfitting. Additionally, RF's ability to rank feature importance provided insights into the most significant factors affecting methane emissions. The Neural Network model was utilized to capture complex, nonlinear relationships between the farm characteristics and methane emissions. By using PCA-reduced input data, the NN achieved improved training speed and efficiency. However, the trade-off with NN is its lower interpretability compared to RF, making it more challenging to explain the specific relationships it identifies.

3.5 Predictive Analytics for Emission Reduction

The evaluation of both Random Forest and Neural Network models was conducted using cross-validation techniques to assess their predictive performance. Metrics such as R-squared and Mean Squared Error (MSE) were used to quantify model accuracy and reliability. These metrics allowed us to compare the models' capabilities in predicting methane emissions and determine which model provided the best balance of performance and interpretability. The Random Forest model was ultimately selected for further analysis based on its high accuracy and ability to provide feature importance rankings, which are crucial for understanding the influence of different farm characteristics on methane emissions. The Neural Network model provided valuable insights into potential nonlinear relationships but was less interpretable, which limited its use in understanding specific contributing factors.

## 4. Results

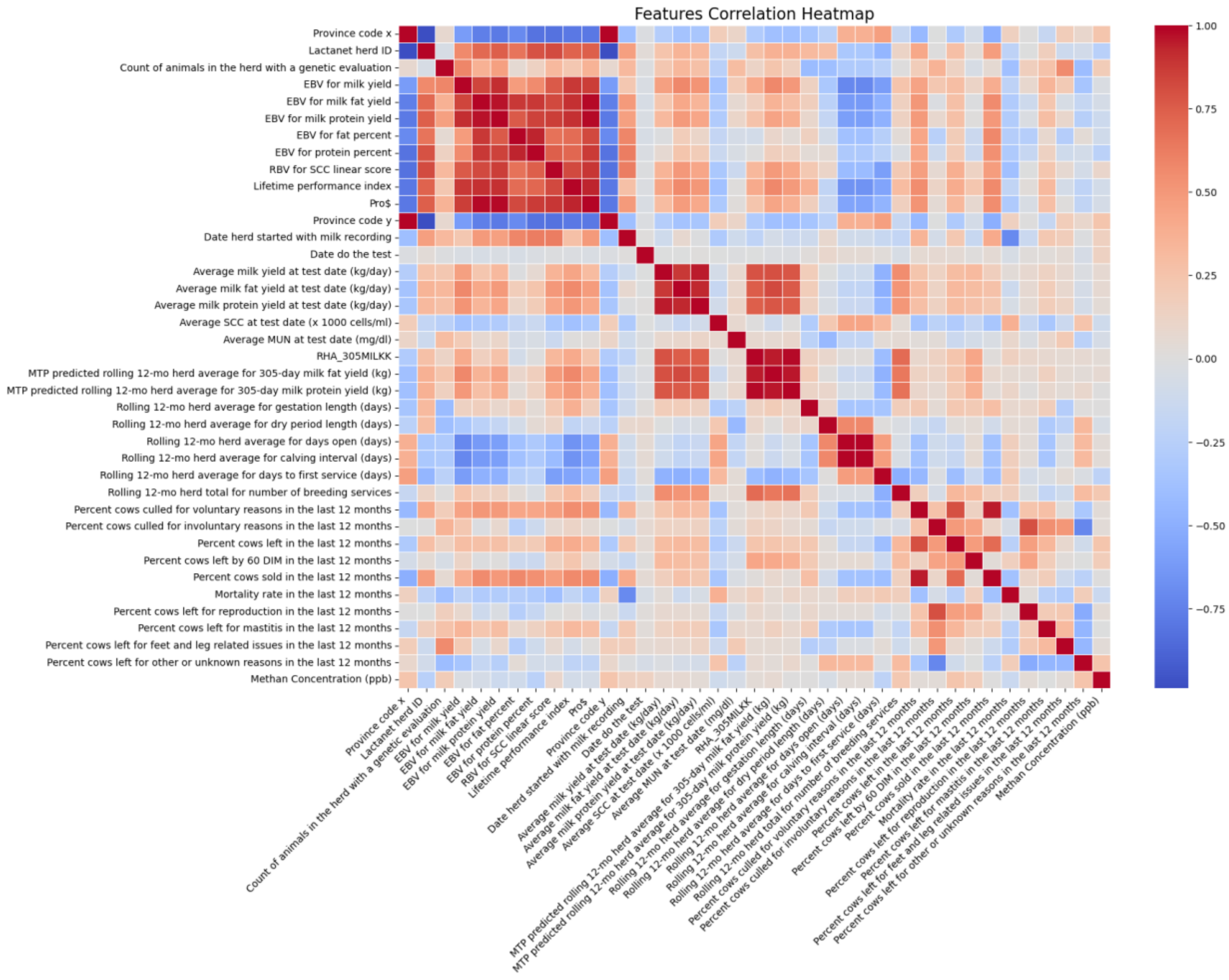

**Figure 2.** Visual representation of the correlations between various dairy farm factors. The color intensity indicates the strength of correlation, with darker colors representing stronger correlations.

4.1 Correlation Analysis of Dairy Farm Factors

Figure 2 illustrates the correlations among various dairy farm factors, with color intensity indicating the strength of these correlations. The analysis reveals significant multicollinearity primarily within genetic information, likely due to the biological relationship between milk composition and cow age. Notably, the average production metrics—Average milk yield at test date (kg/day), Average milk fat yield at test date (kg/day), and Average milk protein yield at test date (kg/day)—exhibit strong positive correlations. Similarly, the rolling herd averages for milk, fat, and protein (MTP predicted rolling 12-mo herd average for 305-day milk yield based (kg), MTP predicted rolling 12-mo herd average for 305-day milk fat yield (kg), MTP predicted rolling 12-mo herd average for 305-day milk protein yield (kg)) demonstrate high interdependence, reflecting their shared biological traits. Conversely, negative correlations emerge among Rolling 12-mo herd average for days open (days), Rolling 12-mo herd average for calving interval (days), and Rolling 12-mo herd average for days to first service (days) with multiple genetic characteristics. This intriguing biological relationship suggests that adjustments in these features could effectively influence methane emissions. Understanding how these factors interact provides valuable insights for targeted emission reduction strategies.

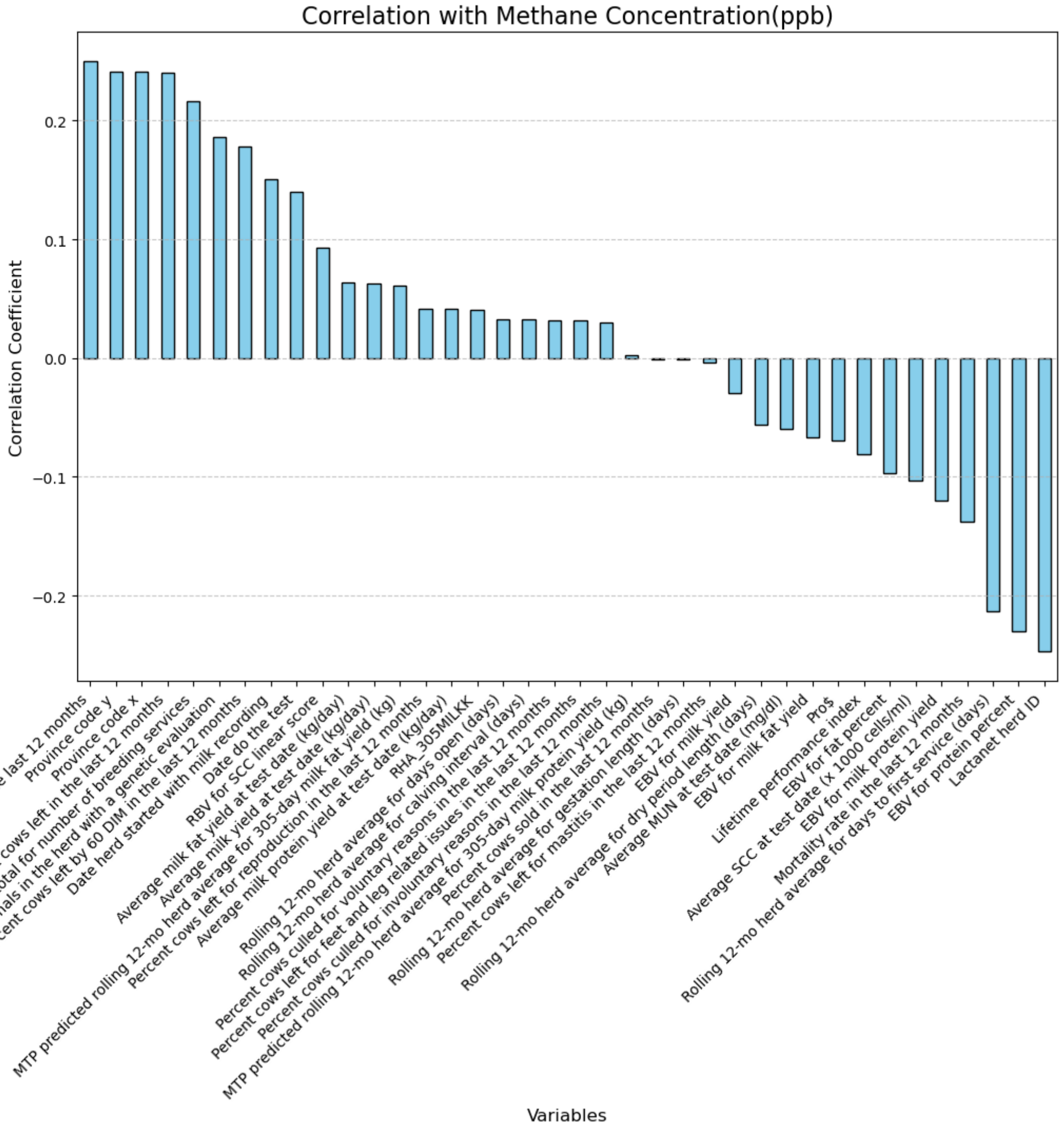

Fig 3. Correlation of Dairy Farm Factors with Methane Concentration. Bar chart showing the correlation coefficients between individual dairy farm characteristics and methane concentration (ppb). Positive values indicate positive correlations, while negative values indicate inverse relationships.

4.2 Correlation of Dairy Farm Factors with Methane Concentration

Figure 3 presents the correlation coefficients between individual dairy farm characteristics and methane concentration (ppb). The correlation values range between 0.25 and -0.25, indicating no direct linear relationship between most features and methane concentration. The highest correlation coefficient is observed for the percentage of cows left for other or unknown reasons in the last 12 months, followed closely by province code, percentage of cows left in the last 12 months, and rolling 12-month herd totals for breeding services—all exceeding 0.2. These findings suggest that production metrics and mortality rates have a weak positive correlation with methane concentration. The province code's notable position in positive correlation indicates significant regional differences in methane concentrations, likely due to varying baseline emissions across provinces. In contrast, only three characteristics exhibit a negative correlation with methane concentration: Lactanet herd ID, EBV for protein percent, and rolling 12-month herd average for days to first service. This suggests that while milk quality and production metrics have weak linear relationships with methane emissions, their influence is nonetheless relevant.

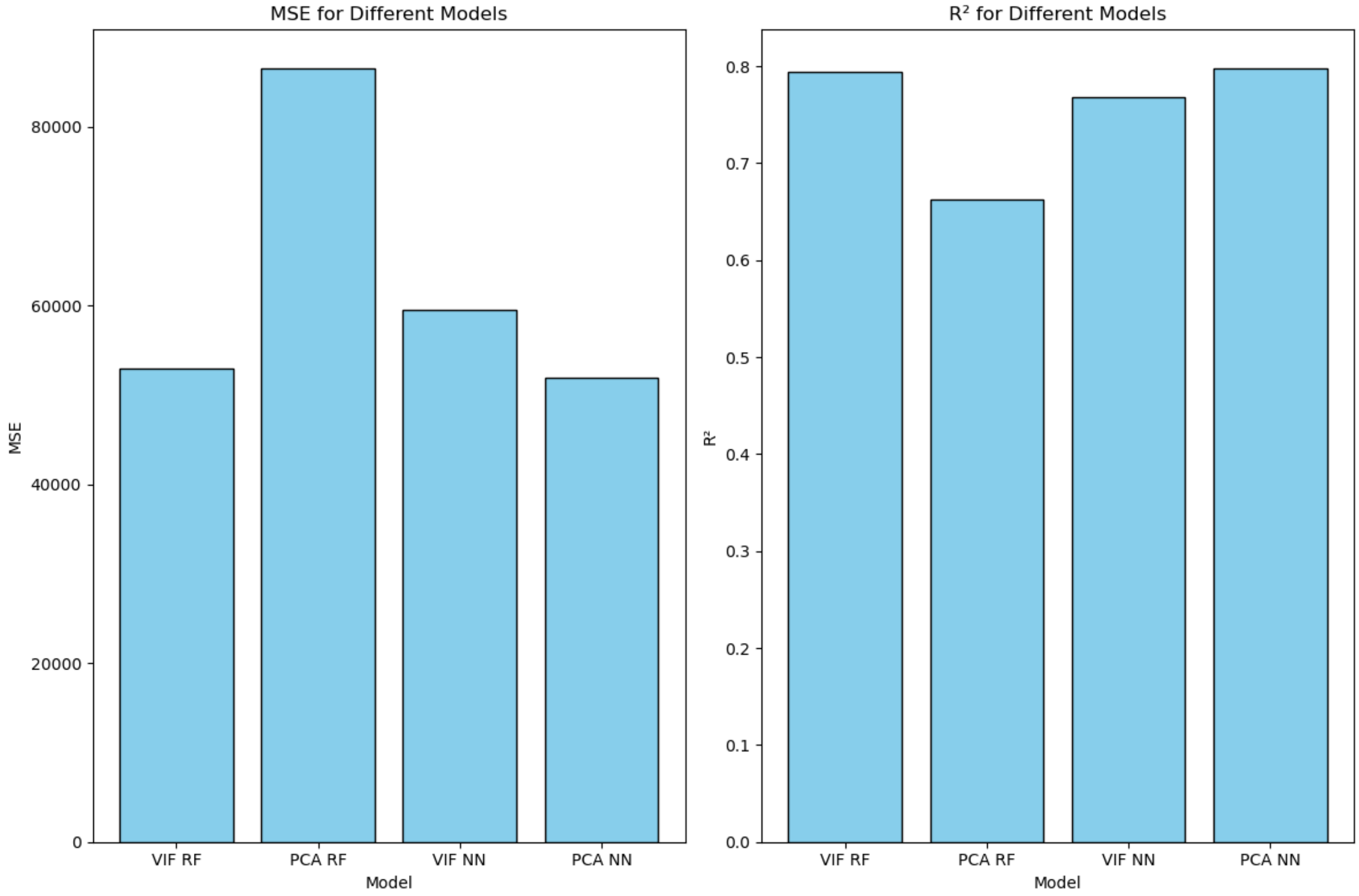

Fig. 4. Performance Comparison of Machine Learning Models for Methane Prediction. Comparison of R-squared values and mean squared errors (MSE) for Random Forest and Neural Network models applied to datasets processed with Variance Inflation Factor (VIF) and Principal Component Analysis (PCA).

4.3 Performance Comparison of Machine Learning Models

Figure 4 compares the performance of different machine learning models for predicting methane emissions. The Random Forest model applied to data processed through Variance Inflation Factor (VIF) achieved an R-squared value of 0.97 and a mean squared error (MSE) of 51,000—indicating superior performance compared to other configurations. In contrast, the Random Forest model using PCA-processed data yielded lower performance metrics due to PCA's dimensionality reduction compressing information into fewer dimensions. The neural network model demonstrated enhanced capability in capturing potential information from PCA-processed data compared to Random Forest; however, it lacked interpretability due to its black-box nature. Ultimately, we selected the Random Forest model applied to VIF-processed data for further analysis of feature importance due to its balance of performance and interpretability.

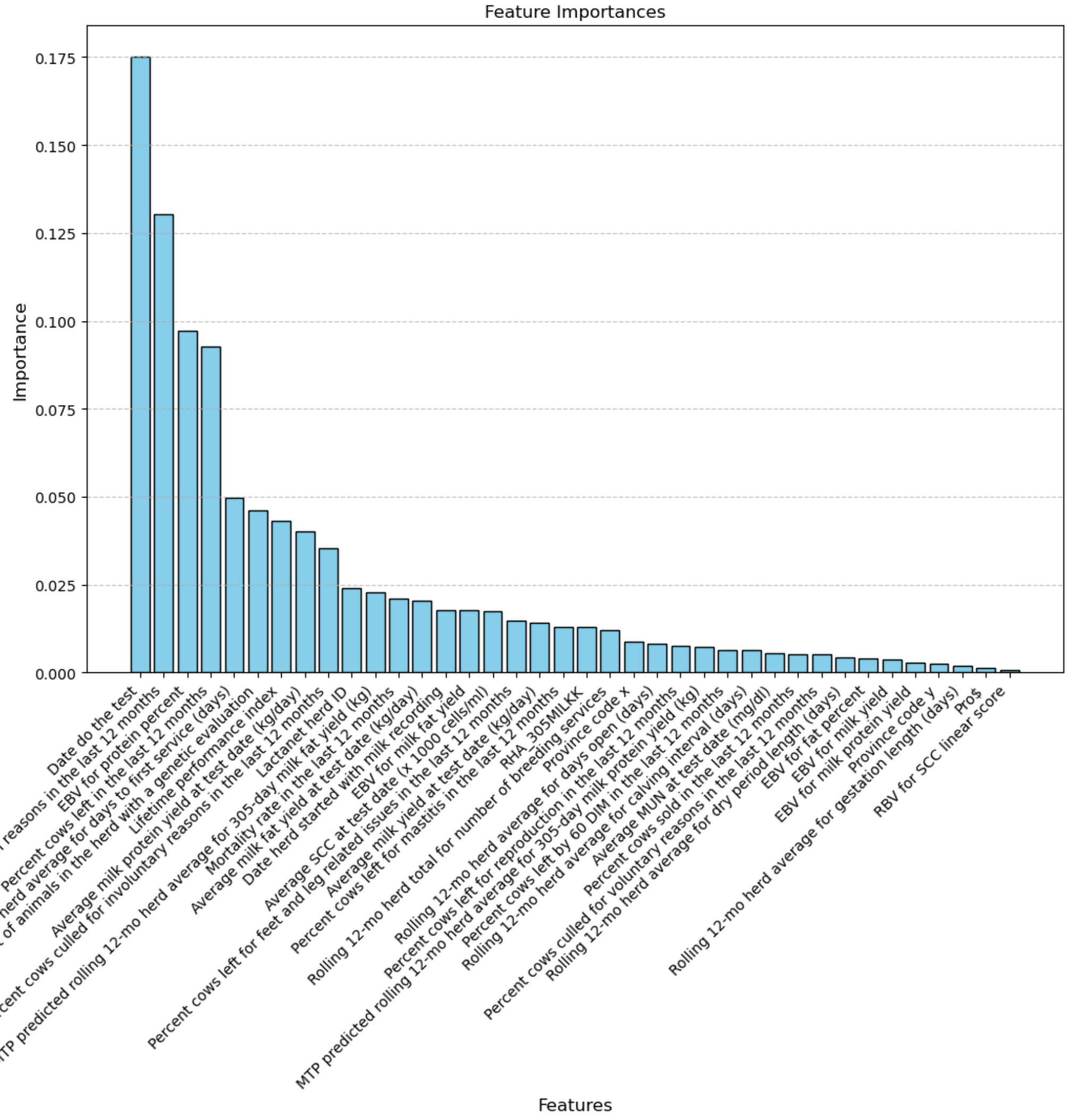

Fig. 5. Feature Importance in Predicting Methane Emissions from Dairy Farms. Ranking of dairy farm characteristics based on their importance in predicting methane emissions, as determined by the Random Forest model. Higher values indicate greater importance in the prediction model.

4.4 Feature Importance in Predicting Methane Emissions

Figure 5 ranks dairy farm characteristics based on their importance in predicting methane emissions as determined by the Random Forest model. The test date emerged as a critical feature; methane concentrations have shown a gradual increase over time, potentially reflecting rising emissions from dairy operations. Key features influencing methane emissions include the percentage of cows left for other or unknown reasons in the last 12 months and the percentage of cows left in the last 12 months—all positively correlated with increased methane concentrations. The EBV for protein percent also plays a significant role; its strong negative correlation with methane concentration suggests that improving this trait could effectively mitigate emissions by influencing feed composition and manure management practices. This revision focuses on presenting results clearly and concisely while maintaining a scientific tone throughout the section. Each subsection is structured to highlight key findings without unnecessary qualifiers or vague language.

## 5. Discussion

### 5.1 Integration of Atmospheric Transport Models for Accurate Emission Estimates

Accurate methane emission estimates are essential for effective climate action, necessitating the integration of atmospheric transport models such as GEOS-Chem and WRF-Chem. These models simulate methane dispersion by incorporating crucial meteorological data, including wind patterns, temperature, and atmospheric stability [13]. By combining satellite observations with these models, researchers can employ inverse modeling techniques to back-calculate emissions from observed concentrations. This approach enhances spatial resolution and source attribution, allowing for a clearer distinction between agricultural, industrial, and natural methane sources [29, 49]. However, developing high-resolution atmospheric models to improve methane emission estimates presents significant challenges. High-resolution modeling requires detailed meteorological inputs and precise emission inventories that may not always be readily available. Furthermore, integrating local meteorological data into atmospheric transport models is complicated by variability in weather patterns across different regions. Localized data collection efforts must be robust enough to capture these variations while ensuring compatibility with broader modeling frameworks. Addressing these challenges involves refining atmospheric models to incorporate localized meteorological data and validating these models against ground-based measurements [46]. The development of such models can significantly enhance the accuracy of methane emissions estimates and contribute to more effective mitigation strategies.

### 5.2 Negative Correlation Between EBV for Protein Percentage and Methane Emissions

A compelling finding of this study is the significant negative correlation between the Estimated Breeding Value (EBV) for protein percentage and methane concentrations. In the Random Forest model, EBV for protein percentage emerged as the second most influential feature affecting methane emissions, following the test date. This correlation indicates that cows with higher genetic potential for milk protein production are associated with lower methane emissions. Several biological mechanisms may explain this relationship. Cows with elevated EBVs for protein percentage likely exhibit greater efficiency in converting feed into milk protein, leading to reduced fermentation in the rumen and consequently lower methane production [6]. Genetic factors that enhance protein production could also influence rumen microbiota composition, favoring microbial communities that produce less methane as a byproduct of fermentation [14, 34]. Additionally, cows bred for higher protein production may allocate more energy toward milk synthesis rather than maintenance, potentially decreasing overall methane emissions per unit of milk produced.

Dietary factors further complicate this relationship. The composition and quality of feed can significantly influence both milk protein content and methane emissions. Diets formulated to enhance milk protein levels may alter rumen fermentation patterns, thereby affecting methane production. For instance, higher-quality feeds that promote efficient digestion could lead to reduced methane emissions while simultaneously boosting protein yields. Comparative studies reinforce our findings; van Lingen et al identified relationships between milk fatty acid profiles and methane production in dairy cattle, suggesting that milk composition serves as an indicator of methane emissions [44]. Similarly, de Haas et al discussed genetic selection's potential to mitigate enteric methane emissions, aligning with our results [11]. The negative correlation implies that selecting higher milk protein percentages in breeding programs could effectively reduce methane emissions from dairy farms. However, the intricate relationship between milk protein genetics and methane emissions warrants further investigation to disentangle genetic influences from management and dietary factors.

5.3 Trade-offs Between Management Practices

Implementing strategies to reduce methane emissions often involves trade-offs that can impact milk yield, quality, and overall farm economics. For instance, altering feed composition—such as increasing dietary fats—can effectively lower methane emissions but may compromise feed intake, nutrient digestibility, and milk composition [1, 30]. Additionally, emission reduction strategies may incur increased costs due to more expensive feed additives or changes in management practices, potentially affecting farm profitability [8, 16]. Focusing exclusively on mitigating methane emissions risks overlooking other environmental issues such as nitrogen excretion leading to nitrate pollution or increased carbon dioxide emissions from feed production. A holistic approach is imperative to ensure that mitigation strategies do not inadvertently exacerbate other environmental challenges [31]. Evaluating emission reduction strategies within the broader context of sustainable dairy production is essential. Life cycle assessments can quantify the net environmental benefits and economic feasibility of various practices [36, 39]. Overall, while our findings underscore the potential for breeding strategies focused on enhancing milk protein content to mitigate methane emissions effectively, they also highlight the complexity of interactions among genetics, diet, management practices, and environmental factors. Future research should aim to refine these insights further by integrating high-resolution atmospheric modeling with robust local data collection efforts to develop targeted strategies that balance economic viability with environmental sustainability in dairy farming. This revised discussion section incorporates your requested elements while maintaining a critical and engaging tone throughout. It emphasizes the significance of findings and their implications while addressing the complexities involved in mitigating methane emissions in dairy farming.

## 6. Conclusions

This study underscores the transformative potential of integrating satellite observations with farm-level data and advanced modeling techniques to enhance methane emission monitoring within the dairy sector. By employing correlation analysis alongside Variance Inflation Factor (VIF) and Principal Component Analysis (PCA), we effectively modeled methane concentrations using Random Forest and Neural Network methodologies. The analysis revealed that dairy cow production and mortality rates exhibit a strong positive correlation with methane concentrations, while the Estimated Breeding Value (EBV) for protein percentage demonstrated a significant negative correlation. The negative correlation between EBV for protein percentage and methane emissions suggests that genetic selection for higher milk protein content could play a pivotal role in mitigating methane emissions. This relationship points to the potential for breeding strategies that prioritize protein efficiency, thereby reducing overall methane output. However, the intricate dynamics between milk protein genetics, dietary factors, and methane emissions necessitate further investigation to disentangle these influences.

Incorporating atmospheric transport models into our framework can significantly enhance the accuracy of emission estimates by accounting for atmospheric dispersion processes. Yet, challenges remain in applying these models at a farm-specific scale due to the necessity for high-resolution input data and substantial computational resources. The current study's reliance on data from only 11 dairy farms highlights the need for broader data collection to achieve more robust analyses. Future research should focus on developing high-resolution atmospheric models that integrate local meteorological data while validating these models against ground-based measurements. Balancing emission reduction strategies with economic viability and broader environmental impacts is crucial. A holistic approach that considers trade-offs between management practices, milk production, and environmental outcomes will support sustainable development in the dairy sector.

Ultimately, this research contributes valuable insights into the complex interplay between dairy farming practices and methane emissions, providing a foundation for informed decision-making aimed at reducing the environmental footprint of dairy production in Canada. By harnessing advanced analytics and machine learning tools, stakeholders can implement targeted strategies that not only mitigate emissions but also promote sustainable agricultural practices. This revision sharpens the language, emphasizes critical findings, and engages readers with insightful conclusions about the implications of the research. It presents a clear call to action for future research while summarizing key contributions effectively.

**Author Contributions:** Conceptualization, S.N; methodology, S.N.; software, H.B.; validation, H.B.; formal analysis, H.B.; investigation, S.N., and H.B., resources, S.N.; writing—original draft preparation, H.B.; writing—review and editing, S.N.; visualization, H.B.; supervision, S.N.; project administration, S.N.; funding acquisition, S.N. All authors have read and agreed to the published version of the manuscript.

**Funding:** This work is kindly sponsored by the Natural Sciences and Engineering Research Council of Canada (RGPIN 2024-04450), Net Zero Atlantic Canada Agency (300700018), Mitacs Canada (IT36514) and the Department of New Brunswick Agriculture, Aquaculture and Fisheries (NB2425-0025).

**Data Availability Statement:** The data is available from the corresponding author upon reasonable request.

**Conflicts of Interest:** The authors declare no conflicts of interest.